%% file: SpyOrNot.tex
\relax
\documentclass[letterpaper]{article} 
\usepackage{spconf}  
\usepackage{times}  
\usepackage{helvet}  
\usepackage{courier}  
\usepackage{url}  
\usepackage{graphicx}  
\usepackage{soul}	
\usepackage{xcolor} 
\usepackage{tabularx}
\usepackage{multirow}
\usepackage{esvect}
\usepackage{bm}
\usepackage[linesnumbered,ruled]{algorithm2e}

\def\spyornot{{\textsf{LiarOrNot}}}
\def\resistance{{\textsf{Resistance}}}
\def\mafia{{\textsf{Mafia}}}
\def\werewolf{{\textsf{Werewolf}}}
\def\spyrank{{\textsf{LiarRank}}}

\newcommand{\nop}[1]{{}}

\begin{document}\sloppy
%
\title{Automatic Long-Term Deception Detection in Group Interaction Videos}
\name{
	 \parbox{0.84\linewidth}{\centering
	Chongyang Bai$^1$, Maksim Bolonkin$^1$, Judee Burgoon$^3$, Chao Chen$^1$, Norah Dunbar$^4$, Bharat Singh$^2$, V. S. Subrahmanian$^1$, Zhe Wu$^2$}
}
\address{$^1$Dartmouth College, $^2$Univerity of Maryland,\\ $^3$University of Arizona, $^4$University of California Santa Barbara}

\maketitle
\begin{abstract}

\renewcommand{\thefootnote}{\fnsymbol{footnote}}
\footnotetext[0]{Authors' emails: Chongyang Bai (cy@cs.dartmouth.edu), Maksim Bolonkin (mbolonkin@cs.dartmouth.edu), Judee Burgoon (judee@email.arizona.edu), Chao Chen (chaochen3333@gmail.com), Norah Dunbar (ndunbar@comm.ucsb.edu), Bharat Singh (bharat@cs.umd.edu), V. S. Subrahmanian (vs@dartmouth.edu), Zhe Wu (zhewu@umd.edu).}

Most work on automated deception detection (ADD) in video has two restrictions: (i) it focuses on a video of one person, and (ii) it focuses on a single act of deception in a one or two minute video. In this paper, we propose a new ADD framework which captures long term deception in a group setting. We study deception in the well-known Resistance game (like Mafia and Werewolf) which consists of 5-8 players of whom 2-3 are spies. Spies are deceptive throughout the game (typically 30-65 minutes) to keep their identity hidden. We develop an ensemble predictive model to identify spies in Resistance videos. We show that features from low-level and high-level video analysis are insufficient, but when combined with a new class of features that we call \spyrank, produce the best results. We achieve AUCs of over 0.70 in a fully automated setting.
\end{abstract}

\section{Introduction}
Sales presentations, business negotiations and diplomatic talks often involve consistent deception in a group setting. During a sales presentation, the seller may present deceptive information about his products. During nuclear negotiations, a country may be deceptive about its intentions. In particular, deceivers in such situations engage in a mix of truthful and deceptive acts over an extended period of time (anywhere from 30-minutes to days). We focus on group settings in which there is visibility of each participant's face.

Past work on automated deception in video \cite{zhang2007real,PerezRosas15,Wu2017} focuses on videos of a single person in a short (15-200 secs) clip.  In contrast, we present a fully automated system (\spyornot) in which we take a frontal video of a subject interacting with a group and predict whether that person is being deceptive in the long term, i.e. across the duration of a 30-65 minute video. To achieve this, we conducted a study that generated 44 games involving 285 players from 5 sites in 3 countries (Singapore, Israel and the USA) by running a version of the well-known \resistance\ game. \resistance\ and its variants like \mafia\ and \werewolf\ naturally induce long term deception in a highly interactive group setting. \resistance\ usually involves 5-8 players, 2-3 of whom are designated ``spies'' who win the game if they are not discovered. Thus, they must be deceptive throughout the game, but must intermix lies with truth in order to stay undiscovered by others.
We develop methods to predict ``spies'' and ``honest'' players in the game.

In addition to the fact that long-term deception in group settings has been rarely studied, \spyornot\ makes the following innovations. Building on  well-known image (VGG Face) and audio features (Mel-frequency cepstral coefficients),
 (i) we introduce 
a class of histogram-based features that build on well known low-level (eye/head movement, facial action units) and high-level (emotion features from Amazon Rekognition) features. 
(ii) we introduce a novel class of ``meta-features'' called \spyrank\ that builds on the basic features, and (iii) we introduce an ensemble based prediction model. Our 10-fold cross validations split the \emph{entire} set of videos into training and testing sets based on games. Hence, \spyornot\ predicts on games and people that are completely disjoint from those seen in training. We show that \spyornot\ achieves an AUC of 0.705 in this hard test, significantly outperforming other feature classes and past work. Additionally, as our data set was collected across three very different countries and because there may be cultural differences in deception, our results are more robust across cultures than past studies (though much additional work needs to be done to capture African and Latin American cultures as well).

\input{related_work}
\input{scan_dataset}
\input{godds}
\input{experiments}
\input{conclusion}

\bibliographystyle{IEEEbib}
\bibliography{references}
\end{document}

%% file: related_work.tex
\section{Related Work}
Zhang et al. \cite{zhang2007real} were among first to use fine-grained image analysis to detect deception in facial and emotional expressions in static images.
To distinguish genuine facial expressions from simulated ones, they proposed a set of features relying on 58 manually labeled facial points, which makes the approach not fully automated. Michael et al.~\cite{michael2010motion} built upon this approach by proposing a feature called motion patterns, incorporating both head/hand movement and automatic facial landmarks tracking.
The experimental setting in their work, however, was constrained to an interview. \spyornot\ is designed to detect deception in an hour-long group interaction, instead of an interview. Wu et al.~\cite{Wu2017} took advantage of the multi-modal nature of videos to detect deception in courtroom trial videos. They used motion, audio, and text features as well as facial micro-expressions to build a fully automated deception detection engine achieving $0.877$ AUC with inferred micro-expressions. 
This work was tested only for short courtroom videos (which is similar to an interview) and not under group interactions.

Chittaranjan et al.~\cite{Chittaranjan2010} pioneered the approach of using videos of games. They collected a dataset of \werewolf\ videos which is similar to \resistance. They used verbal and non-verbal cues to predict players considered deceitful by other players. They did not take visual appearance into account. Moreover, they focused on predicting other players' perception of deceitful behavior rather than actually predicting the werewolves (who are similar to the spies in \resistance). Demyanov et al.~\cite{Demyanov2015} created a dataset of \mafia\ game videos and proposed a method to detect deceptive players. They achieved $0.639$ AUC by analyzing facial action units of players.
Yu et al.~\cite{Yu2015}'s important paper considered a game called ``Killer Game" with a similar set up. In this study players participated in the game online via voice or text messages. Yu et al.~\cite{Yu2015} used sentiment analysis to infer players' attitude towards each other and to build a network to identify a group of deceitful players.

Unlike previous studies~\cite{zhang2007real, Wu2017}, we deal not with short videos of an act of deception but rather with long (30--65 minutes) videos of humans, some of whom are actively avoiding being deceptive. Thus, it is impossible to select a specific point in time when deception is happening, and the decision whether a player is a spy or a member of resistance should come from analyzing the whole video. Unlike ~\cite{Yu2015}, we actively use audio--visual information; we ignore transcript analysis for now.
We build on the use of Facial Action Units as in ~\cite{Demyanov2015}. In addition, we use emotion predictions provided by Amazon Rekognition, as well as some low level features such as eye/head movements and Convolutional Neural Network representations. Additionally, we propose a new class of meta-features called \spyrank.

%% file: scan_dataset.tex
\section{Game and Dataset description}
\label{sec:dataset}
Our \resistance\ dataset contains a set of videos depicting groups of 5-8 people playing a social game.

\emph{The Game.}
Each player is secretly told that she belongs to a team of ``spies'' or a team of ``resistance''. Spies know who other spies are, but the resistance does not know any information. There are 2--3 spies in a game. The game proceeds in rounds (typically 3 to 7 in a game) called missions. 
Every round has three stages: players nominate and elect a mission team leader; the leader nominates mission team members, and players vote for that mission team; finally, the mission team ``goes on the mission''.
In the leader nomination stage, players get nominated to serve as a leader. All players vote for or against the nominee. This stage is repeated until the team leader is elected. In the second stage of the round, the team leader nominates team members. After a discussion, all players vote on approval or rejection of the proposed team. This stage is repeated up to three times or until the team is approved.
In the third stage the team members secretly vote for the success or failure of the mission. Spies want the mission to fail, resistance want the mission to succeed.
If the vote is in favor of mission success, the resistance team collectively gets a point. If some votes go the other way, the spies collectively get a point. Spies also score a point if players fail to approve the proposed team three times. A team (spies or resistance) with the highest score at the end of the game wins. Therefore spies have a natural incentive to get elected as team leaders and to get on mission teams. For the resistance team it is advantageous to identify spies as soon as possible and prevent them from getting on mission teams, which means spies need to make sure they are not discovered.

\emph{Dataset description}. Our \resistance\ dataset contains a set of videos of \resistance\ games involving 285 players (total of 113 spies and 172 members of resistance) collected from 5 sites spread over 3 countries (three locations in USA plus Israel and Singapore). Videos span a minimum of 30 minutes to a maximum of 65 minutes with the average duration being 46 minutes.
In this paper we use video of a player captured by a tablet camera directly in front of the player. Since the players were interacting continuously throughout the video, each camera also captured audio of all players.
\vspace{-3mm}

%% file: godds.tex
\begin{figure*}[ht]
	\centering
		\includegraphics[width=\textwidth,scale=1]{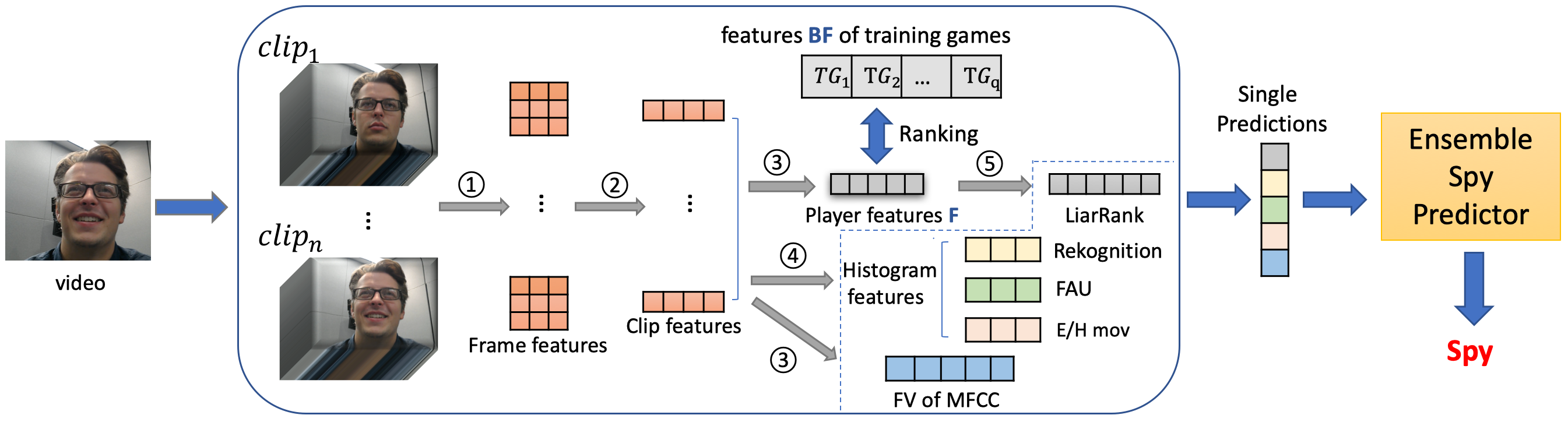}
\caption{\spyornot\ Architecture. Steps: Uniformly sample n clips from a player's video, then (1) extract frame features, including VGG Face, emotions, facial action units and eye/head movements, (2) aggregate frame features and sub-second MFCC features to clip features, (3) and (4) aggregate previous features to player features,  (4) is histograms of low-dimension high-level features, and (3) is Fisher Vectors (FV), (5) build \spyrank\ of player features. Finally, predictions made from each feature type are used in our ensemble spy predictor to generate the final prediction.}
	\label{fig:framework}
\end{figure*}

\section{\spyornot\ Deception Detection System}
\noindent\emph{Architecture.} Figure~\ref{fig:framework} shows the \spyornot\ architecture. Let $\mathcal{TG}=\{TG_1,\ldots,TG_n\}$ be the set of training game videos (e.g. in some fold of cross validation) and 
let $TG_{n+1}$ be any game (either in $\mathcal{TG}$ or not). In any game
 $TG_j$, let $p_i^j$ be the $i$'th player in that game. In our data,  $i$ varies from 1 through a max of 8. Each player $p^i_j$'s frontal camera captures a video $v^i_j$ of that player of length 30--65 minutes. \emph{Each player appeared in exactly one game.} Since we wish to predict whether a player $p^i_j$ is deceptive or not, each player needs to have an associated feature vector $fv(p^i_j)$ which we define as either a basic feature vector $bf(p^i_j)$ or a \spyrank\ meta-feature vector $sr(p^i_j)$.\\
The rest of this section is organized as follows. We first explain the concept of \spyrank, showing how to associate a \spyrank\ meta-feature vector $sr(p^i_j)$ with player $p^i_j$. We then explain how the ``basic'' features are derived. Finally, we explain our ensemble predictor. Throughout this section, we use the "dot" notation to denote the connection between representations and level of aggregation, e.g. $fr.f_i$ denotes feature $f_i$ of the frame $fr$, and $Cl.\bm{f}$ denotes feature vector $\bm{f}$ of clip $Cl$. 

\vspace{-2mm}
\subsection{\spyrank\ Features}
Suppose $BF=\{f_h\}_{h=1}^k$ is any set of basic features.
Given any basic feature $f_h$, we will first define the \spyrank\ $sr_h(p^i_j)$ of player $p^i_j$ w.r.t. feature $f_h$.
The \spyrank\ vector $sr(p^i_j)$ is then the vector 
$\langle sr_1(p^i_j),\ldots ,sr_k(p^i_j)\rangle$ obtained by concatenating these individual feature-ranks.

\begin{algorithm}[t!]
    \caption{\textsc{\spyrank}$(\mathcal{TG},TG_{n+1},p_{n+1}^\ell,f_h)$}
        \label{alg}
	\SetKwInOut{Input}{Input}\SetKwInOut{Output}{Output}
    \Input{Training set $\mathcal{TG}=\{ TG_1,\ldots, TG_n\}$, Player $p_{n+1}^\ell$ from some game $TG_{n+1}$, basic feature $f_h$} 
    \Output{$sr_h(p_{n+1}^\ell)$}
    \BlankLine
    	\For{$j \in [1, \ldots ,n]$} { 
	   		{$Vals(f_h,j)= \{p_{n+1}^\ell.f_h\}\cup \bigcup_{i=1}^8 \{ p^i_j.f_h\}$}\\
	   				   {Sort  $Vals(f_h,j)$ in descending order}\\
	   				   {$r_j$= position of $p_{n+1}^\ell.f_h$'s value in $Vals(f_h,j)$}}
 	\Return{the vector $\langle r_1,\ldots,r_n\rangle$}
\end{algorithm}

The \spyrank\ algorithm shown above takes as input, a training set $\mathcal{TG}=\{ TG_1,\ldots, TG_n\}$, a game $TG_{n+1}$ (which could be in $\mathcal{TG}$ or not), as well as a player and a single feature $f_h$. It returns a vector of length $n$ (i.e. number of games in the training set)  which captures the position of players $p_{n+1}^\ell$'s value for feature $f_h$ w.r.t. the corresponding values for other players in each of the $n$ games.
To do this, it computes the value of the feature for the player $p_{n+1}^\ell$ as well as every player who participated in any of the training games. The resulting set of features values is stored in the set $Vals(f_h)$. This set of values is then sorted in descending order. The first item in the descending order has position (or rank) 1, the second has position (or rank) 2, etc. The \spyrank\ of player $p_{n+1}^\ell$ w.r.t. feature $f_h$ is its position in the sorted $Vals(f_h)$ list. \emph{Intuitively, \spyrank\ of player $p_{n+1}^\ell$ w.r.t. feature $f_h$ is the relative rank of player $p_{n+1}^\ell$ had she participated in that game.}

The above defines the \spyrank\ vector of a player w.r.t. a feature. The \spyrank\ vector of a player is the concatenation of the feature vectors. There is some similarity between \spyrank\ and the rank transform proposed in \cite{zabih1994non} and the local binary pattern descriptor (LBP) in computer vision. 

\subsection{Basic Features}
\noindent\emph{Sampling}.
We sample 10-second clips at an interval of 30 seconds per video.
Since games are 30-65 minutes long, different videos may consist of different numbers of clips. From each clip, we further sample a set of $m=300$ frames for Eye/Head Estimations and $m=20$ frames for the rest of visual features (see below). As low/high-level video features as well as audio features for each player may vary substantially over the length of the video, we define features at both the frame-level and clip-level. For each $(ClipId,FrameId)$ pair, we extract a set of basic features.

\noindent\emph{Basic Frame Features.} For sampled frames, we extract the following basic features: VGG Face \cite{Parkhi15}; Facial Action Units (FAU) and Eye/Head Estimations (E/H) using OpenFace ~\cite{8373812}; Amazon Rekognition for 7 emotions (happy, sad, angry, confused, disgusted, surprised, calm) and 3 facial attributes (open eyes, open mouth, smile); and MFCC features~\cite{davis1980comparison}.

\noindent\emph{Basic Clip-level features.} 
We aggregate frame-level features into clip-level features with average-pooling.
If a clip $Cl$ is a set of sampled frames, then the value of a clip-level feature $f_h$
 for clip $Cl$ is given by $Cl.f_h=\frac{1}{|Cl|} \Sigma_{fr\in Cl} fr.f_h$.
 Clip-level features smooth variations in frame level features, especially as those variations can be substantial for some features, e.g. emotion features.

\noindent\emph{Player-level features.} As the goal is to extract features at a per-player level, we aggregate clip-level features into player-level features using Fisher Vectors (for VGG Face representations), or histograms (for Facial Action Units, Eye/Head movement, and Amazon Rekognition features).

\noindent\emph{Fisher Vector features.}
Fisher vector (FV) is a bag-of-words based model heavily used for object recognition in images. Note that each video may have a different number of clips. Fisher Vectors aggregate the clip level features of an arbitrarily long video into a fixed length encoding. 

\noindent\emph{Histogram features.}
We compute three types of histogram features for every basic feature such as Facial Action Units, Eye/Head movement, and Amazon Rekognition features. These are histograms of frame-level features, histograms of clip-level features, and combination of the first two.\\
For a player $Pl$ and a basic frame feature $f_h$, we have a set of all feature values for all frames $\{fr_{st}.f_h\}$, where $fr_{st} \in Cl_t$ and $Cl_t \in Pl$ (or a set of clip-level features $\{Cl_1.f_h, Cl_2.f_h, \ldots, Cl_{|Pl|}.f_h\}$ where $Cl_i \in Pl$.). We build a histogram of frame-level features $\mathcal{V}^{frames}_h = \langle v_h^1, v_h^2, \ldots, v_h^b \rangle$ where $v_h^i$ are frequencies of values $fr_{st}.f_h$ falling into the $i^{th}$ bin, and $b$ is the number of bins (similarly $\mathcal{V}^{clips}_h = \langle v_h^1, v_h^2, \ldots, v_h^b \rangle$ for a histogram of clip-level features). We form a histogram feature by concatenating histograms for all or some of basic features $Pl.\bm{f} = \langle \mathcal{V}^{frames}_{h_1}, \mathcal{V}^{frames}_{h_2}, \ldots \rangle$ (or $Pl.\bm{f} = \langle \mathcal{V}^{clips}_{h_1}, \mathcal{V}^{clips}_{h_2}, \ldots \rangle$ for clip-level histograms).
Finally, we also build combined histogram features by concatenating frame-level histograms and clip-level histograms of the same combination of features $Pl.\bm{f} = \langle \mathcal{V}^{frames}_{h_1}, \mathcal{V}^{clips}_{h_1}, \mathcal{V}^{frames}_{h_2}, \mathcal{V}^{clips}_{h_2}, \ldots \rangle$. Optimal number of bins $b$ is determined through cross-validation.

\subsection{Ensemble classifier}
\label{sec:ensemble}
The previous steps associate with each player $p^i_j$ a feature vector $fv(p^i_j)$ represented by the basic features or associated \spyrank\ features listed above at the player level (aggregating from frame- and clip-levels as described above).
Thus, there are five types of features: \spyrank\ of Fisher Vector of VGG Face, Facial Action Units, Rekognition Emotions, Eye/Head movement, and MFCC. We trained a suite of classifiers and used them to produce a late fusion model. Each classifier returns a \emph{score} denoting the probability of a subject being a spy. If $S_i$ is the score returned by a classifier for the $i$th feature type for $i\in\{1,\ldots,5\}$, then the final score $S$ is obtained by late fusion of named models:
\[S = \sum_{i=1}^{5}\alpha_i S_i~,\]
where $\sum_{i=1}^{5}\alpha_i = 1$. Late fusion weights $\alpha_i$ are obtained by grid-search and cross-validation. 
For each of the five types of features, we select the best classifier, and combine them as above via late fusion.

%% file: experiments.tex
\begin{table*}[h]
\vspace{-4mm}
\centering
\scalebox{0.8}{
\begin{tabular}{|c|c|c|c|c|c|}
\hline 
\multicolumn{6}{|c|}{Amazon Rekognition} \\ 
\hline 
\multicolumn{2}{|c|}{Frame hist.} & \multicolumn{2}{|c|}{Clip hist.} & \multicolumn{2}{|c|}{Combined} \\
\hline 
Disgusted, Surprised & 0.630 & Smile, Angry, Disgusted & 0.634 & Smile, Angry, Disgusted & \textbf{0.676} \\ 
\hline 
Surprised & 0.622 & Smile , Angry & 0.623 & Smile, Disgusted & 0.647 \\ 
\hline 
Calm & 0.622 & Smile, Disgusted, Calm & 0.618 & Angry & 0.638 \\ 
\hline 
All features& 0.557 & All features & 0.544 & All features & 0.563 \\ 
\hline  
\multicolumn{6}{|c|}{Facial Action Units} \\ 
\hline 
\multicolumn{2}{|c|}{Frame hist.} & \multicolumn{2}{|c|}{Clip hist.} & \multicolumn{2}{|c|}{Combined} \\ 
\hline 
AU07+AU10+AU12  & \textbf{0.621} & AU06+AU14 & 0.609 & AU07+AU09+AU10 & \textbf{0.621} \\ 
\hline 
AU12+AU23+AU25 & 0.614 & AU07+AU09+AU10  & 0.606 & AU07+AU10+AU23 & 0.617 \\ 
\hline 
AU09+AU10+AU12  & 0.612 & AU07+AU14+AU45 & 0.603 & AU12+AU25 & 0.611 \\ 
\hline 
All features & 0.592 & All features& 0.577 & All features & 0.608 \\ 
\hline 
\multicolumn{6}{|c|}{Eye/Head movement} \\ 
\hline 
\multicolumn{2}{|c|}{Frame hist.} & \multicolumn{2}{|c|}{Clip hist.} & \multicolumn{2}{|c|}{Combined} \\ 
\hline 
3+8 & 0.632 & 1+6+8 & \textbf{0.671} & 1+3+4+5+6+8 & 0.643 \\ 
\hline 
3 & 0.624 & 1+6 & 0.642 & 1+3+5+8 & 0.627 \\ 
\hline 
3+7 & 0.615 & 1+3+6+8 & 0.636 & 1+3+5+6+8 & 0.625 \\ 
\hline 
All features& 0.591 & All features & 0.560 & All features & 0.618 \\ 
\hline 
\end{tabular}
}
\caption{Performance (AUC) of histogram based representations: top three subsets and all features for frame-level histograms, clip-level histograms, and combined histograms. In all cases sets of all features perform worse than proper subsets due to excessive noise introduced by irrelevant features. For Action Units numbers refer to FACS~\cite{7540162}. Movement features encoding is the following: 1/2: horizontal/vertical eyes movements, 3-5: Euler angles of head rotations, 6-8: $x, y, z$ head translations.}
\label{tab:highlevelres}
\end{table*}


\section{Experiments}
\vspace{-1mm}
\subsection{Experimental setup}
We use videos of 285 players from 44 games. We split the dataset into 10 folds by games, i.e. all players from a game are in either the training or the testing part of a fold.  Our classifier suite includes: 
k-Nearest Neighbors (KNN), Logistic Regression (LR), Gaussian Naive Bayes (NB), Linear SVM (L-SVM) and Random Forest (RF). 
As a performance metric we report the mean AUC over 10 folds.
\vspace{-1mm}
\subsection{Prediction using single-feature classifiers}
\begin{table}[h]
\resizebox{\linewidth}{!}{%
\begin{tabular}{|l|c|c|c|c|c|}
\hline
Features & RF & L-SVM & NB & LR & KNN \\ \hline
Average VGG Face (baseline) & 0.516 & 0.533 & 0.549 & 0.546 & 0.50 \\
VGG Face clip-level voting & 0.503 & 0.520 & 0.550 & 0.527 & 0.479 \\
FV of VGG Face & 0.468 & 0.573 & 0.502 & 0.584 & 0.502 \\
FV of VGG Face + FS & 0.506 & 0.470 & 0.491 & 0.467 & 0.522 \\
\spyrank\ of FV of VGG Face + FS& 0.639 & 0.647 & \textbf{0.663} & 0.652 & 0.603 \\ \hline
FV of MFCC frame-level &0.606 &0.395 & 0.56&0.608 &0.579 \\
FV of MFCC clip-level &0.586 &0.441 &0.533 &0.579 &0.595 \\ \hline

\end{tabular}}
\caption{Performance (AUC) of different aggregations of visual (VGG Face) and audio (MFCC) representations. Top to bottom: 1. Average pooling of all frames; 2. Clip-level VGG Face features are used to train and test, scores are averaged for player-level inference; 3. Fisher Vector of clip-level VGG Face features; 4. Fisher Vector of clip-level VGG Face features after feature selection procedure; 5. \spyrank\ of the Fisher Vector of clip-level VGG Face features after feature selection; 6. Fisher Vector of all MFCC features;  7. Fisher Vector of clip-level MFCC features.}
\vspace{-2mm}

\label{tab:lowlevres}
\end{table}
\emph{\spyrank}. Table \ref{tab:lowlevres} shows performance of different aggregations from VGG Face-based and MFCC-based features including \spyrank. As a baseline we use the feature obtained by averaging all frame-level VGG Face features. This baseline does not even achieve 0.55 AUC, which means simple averaging is not a good strategy to capture the relevant behavior of a player over a long video.\\
Another baseline we explore is to consider every clip-level feature as a point in the dataset, and to assign each clip the label of the player this clip belongs to. To generate player-level predictions, we perform inference for every clip and average clip-level predictions. The highest AUC we achieve using VGG Face is 0.55, which supports the claim that for deceptive behavior detection it is necessary to consider video as a whole.\\
Fisher Vector (FV) is better than the above baselines, achieving an AUC of 0.584. We attribute this to the fact that FV captures statistical information from the whole video rather than from a short clip.\\
Finally, \spyrank\ of Fisher Vector of VGG Face feature obtains the highest 0.663 AUC after feature selection (FS), and this improvement is statistically significant ($p < 0.01$). To verify that improvement comes from the proposed meta-feature and not merely from feature selection procedure, we perform feature selection on Fisher Vector of VGG Face (base feature for \spyrank\ in our experiments), which achieves the highest AUC of 0.522. This experiment suggests that \spyrank\ is important for the improvement in accuracy.
\emph{Histogram features}. As baselines we use mean values of Amazon Rekognition features, Facial Action Units and Eye/Head movement features over all the frames in a video. Although some of these baselines (0.586 AUC for Amazon Rekognition, 0.6 AUC for Facial Action Units and 0.5 AUC for Eye/Head movement) outperform VGG Face baselines, they are significantly inferior to histogram-based player-level features based on corresponding frame-level features.\\
Each aforementioned frame-level representation consists of several features corresponding to individual emotions or facial expressions, not all of which are useful for the task of deception detection. To address this problem, we perform cross-validation with exhaustive search through all possible combinations of features within every representation. So, when computing histogram vectors, we concatenate histograms of a subset of features.\\
Table \ref{tab:highlevelres} shows that different ways of producing histograms (from frame-level features and from clip-level features) perform differently not just in terms of classification performance but also in terms of best subset of features. In case of Amazon Rekognition features and Facial Action Units, it is advantageous to use combined histogram features. For Eye/Head movement features, however, clip-level histograms yield the best performance.\\
Our experiments show that for Amazon Rekognition based features, the combination of three expressions ``Smile", ``Angry" and ``Disgusted" performs the best and achieves 0.676 AUC. For Facial Action Units, the combination of AU07, AU09 and AU10 
achieves 0.621 AUC. The combination of horizontal eyes movements and $x,z$ head translations achieves 0.671 AUC. 
In all cases representations including all the individual feature histograms ("All features" in Table \ref{tab:highlevelres}) perform worse than some of the subsets.\\

\vspace{-3mm}
\subsection{Ensemble Prediction and Feature Importance}
For our ensemble classifier, we use five best performing features: histogram features of facial action units (AU07, AU09, AU10), Fisher Vectors of MFCC, histogram features of Amazon Rekognition predictions (Smile, Angry, Disgusted), histogram features of best movement feature combinations in Table \ref{tab:highlevelres} and \spyrank\ of VGG Face Fisher Vector. Since for single-feature experiments we use a number of classifiers, we perform exhaustive search through all possible combinations of classifiers for the mentioned features. Once single-feature classifiers are trained, we perform late fusion using grid search as described in the Section~\ref{sec:ensemble}. Table \ref{tab:ensemble} shows our Top-5 ensemble prediction results, including what classifiers were used for the corresponding features. Best predictive models yield an AUC of 0.705.\\
To assess the  importance of features for the ensemble classifier, we repeated the process leaving out one class of features at a time. We show the results of this ablation experiment in Table \ref{tab:leaveoneout}. We can see that \spyrank\ of VGG Face Fisher Vectors and the Emotion (Amazon Rekognition) histogram features are the most important.

\begin{table}[h]
\resizebox{\linewidth}{!}{%
\begin{tabular}{|l|c|c|c|c|c|c|}
\hline
Classifiers & AUC & F1 & FNR & FPR & Precision & Recall\\ \hline 
LR+RF+NB+L-SVM+NB & \textbf{0.705} &	0.466 &	0.621 &	0.142 &	0.666 &	0.379 \\ \hline
LR+L-SVM+NB+L-SVM+NB & \textbf{0.705} &	0.466 &	0.610 &	0.169 &	0.660 &	0.390 \\ \hline
KNN+RF+NB+RF+NB & 0.704 &	0.403 &	0.673 &	0.173 &	0.622 &	0.327 \\ \hline
NB+L-SVM+NB+L-SVM+NB & 0.704 &	0.406 &	0.667 &	0.151 &	0.624 &	0.333 \\ \hline
LR+KNN+NB+L-SVM+NB & 0.704 &	0.468 &	0.620 &	0.143 &	0.684 &	0.380 \\ \hline
\end{tabular}}
\caption{Performance (AUC) of Top 5 ensemble models. Classifiers in the table are trained on the features in the following order: histograms of AU07, AU09, AU10; Fisher Vectors of MFCC; histograms of Smile, Angry, Disgusted; histograms of horizontal eyes movement, $x$ and $z$ head movement; \spyrank\ of VGG Face Fisher Vector.}
\label{tab:ensemble}
\end{table}

\begin{table}[h!]
\resizebox{\linewidth}{!}{%
\begin{tabular}{|l|c|c|c|c|c|c|}
\hline
Removed feature & AUC & F1 & FNR & FPR & Precision & Recall\\ \hline 
MFCC & 0.703 &	0.463 &	0.610 &	0.175 &	0.655 &	0.390 \\ \hline
E/H Movement & 0.703 &	0.508 &	0.548 &	0.197 &	0.599 &	0.452 \\ \hline
FAUs & 0.702 &	0.448 &	0.598 &	0.209 &	0.587 &	0.402 \\ \hline
Amazon Rek. & \textbf{0.688} &	0.524 &	0.485 &	0.281 &	0.556 &	0.516 \\ \hline
\spyrank & \textbf{0.688} &	0.411 &	0.344 &	0.721 &	0.104 &	0.560 \\ \hline
\end{tabular}}
\caption{Classification performance (AUC) when one feature class is left out in ensemble predictions. Features details are in Table \ref{tab:ensemble}.}
\label{tab:leaveoneout}
\vspace{-2mm}
\end{table}

\subsection{Human Study}
To assess the complexity of the task and obtain some objective baseline we conducted a human study using the Amazon Mechanical Turk service. To provide a fair comparison, we presented workers with the same data we are using for testing our model: we stitched 10-second clips together with a 1 second transition between them keeping the sound on. Workers were provided with a brief description of the game they were about to watch and asked to make a decision whether the player in the video was a spy or a member of the resistance. To further verify the quality of annotations, workers were asked to provide written justification for their decision.\\
We selected 10 games containing 66 videos in total, and got every video annotated by 3 different workers. Correct player's role was guessed by a majority (2-3 workers out of 3) only in 53\% of videos. We also used the average vote of turkers as a prediction score for the video. In this case, the AUC for human prediction is 0.583, while our ensemble predictor gets 0.701 AUC for the same data ($p < 0.01$). This suggests that detecting deception in long videos is a hard task for humans. We also found that in more than 80\% of the videos, players were suspected to be spies when the actual ratio of spies in the dataset was 42\%. This means that humans, when presented with the fact that a player could be a spy, tend to interpret a player's behavior as suspicious.

%% file: conclusion.tex
\section{Conclusions}
\vspace{-1mm}
We presented an ensemble based automated deception detection framework called \spyornot\, which predicts deception in a group setting by processing long videos. Our framework utilizes appropriate representations at different temporal resolutions for multiple features which capture low and high level information. We also propose a novel class of meta-features called \spyrank\, which provides a significant boost in overall performance. We evaluated \spyornot\ on a dataset collected across different sites and cultures. In a rigorous cross-validation based testing protocol, which separates identities and games during training and inference, we obtained an AUC greater than 0.7, which was 12\% better than average human performance.

\noindent\emph{Role of Authors.} Authors Burgoon and Dunbar designed the \resistance-style game, designed how the game would be run face to face, and collected the \resistance\ data.
The remaining authors designed the feature extraction and machine learning algorithms and software, and designed/ran all experiments.

\noindent \textbf{Acknowledgement.} This work was funded by ARO Grant W911NF1610342.